\documentclass[preprint,12pt]{elsarticle}

\usepackage{amssymb}
\usepackage{float}
\usepackage{url}
\usepackage{graphicx}
\usepackage{amsmath}
\DeclareMathOperator*{\argmin}{arg\,min}
\journal{Pattern Recognition}

\begin{document}
	
	\begin{frontmatter}
		
		\title{Spectral Collaborative Representation based Classification for Hand Gestures recognition on Electromyography Signals}
		
		\author{Ali BOYALI}
		\address{National Institute of Advanced Industrial Science and Technology \\ 1-1-1 Umezono, Tsukuba, 305-8568, Ibaraki, Japan}

		\begin{abstract}
			
			In this study, we introduce a novel variant and application of the Collaborative Representation based Classification	in spectral domain for recognition of the hand gestures using the raw surface Electromyography signals. The intuitive use of spectral features are explained via circulant matrices. The proposed Spectral Collaborative Representation based Classification (SCRC) is able to recognize gestures with higher levels of accuracy for a fairly rich gesture set. The worst recognition result which is the best in the literature is obtained as 97.3\% among the four sets of the experiments for each hand gestures. The recognition results are reported with a substantial number of experiments and labeling computation.  
			
		\end{abstract}
		
		\begin{keyword}
			\texttt{EMG gesture, continous gesture recognition, spectral representation, gesture training matrix, MYO armband}
			\MSC[2010] 00-01\sep  99-00
		\end{keyword}
		
	\end{frontmatter}

	\section{Introduction}
	
	Electromyographic (EMG) signal based applications beyond clinical and rehabilitation purposes have been extensively studied not only to analyze and improve the sport \cite{clarys1993electromyography, masso2010surface},  occupational \cite{clarys2000electromyography} and art performance \cite{shan2012emg} in which kinesiological EMG signals are captured via surface EMG electrodes but also to create intuitive Human Machine Interfaces (HCIs).  
	
	In the latter group, the control of exoskeletons, robotic prosthetic arms and hands are the high end applications that require classification of the EMG signal patterns acquired from the related muscles groups non-invasively.  A vast variety of classification methods have been proposed in the literature.  Among the suggested classification schemes, the heuristic methods such as  the variants of artificial neural networks and fuzzy classifiers \cite{haselsteiner2000using, kale2009intelligent, del1994myoelectric, naik2008multi} prevail among the related studies. Depending on the number of signal patterns to be classified using different kind of  features, methods for pre-processing, source separation and filtering, the accuracy of the classification methods range between 88-99.75\% \cite{ahsan2009emg, chowdhury2013surface}.  
	
	The EMG signal measured on skin surface  is not a mere signal but rather the superposition of the  Motor Unit Action Potential (MUAP) of the many tiny muscle fibers \cite{konrad2005abc, rodriguez2012motor}.   Due to the stochastic nature of MUAP, the EMG signal pattern might exhibit intra and interpersonal variations for the same muscle activity that makes the EMG pattern classification an intricate task for creating robust classifiers.
	
	In concert with the advancements in pervasive computing technologies, the wearable gadgets in which high grade sensors embedded have been released to the market. Thalmic Labs' MYO armband \cite{myoweb} is one of the recent technology which consists of eight surface EMG sensors along with the inertial measurement units with an affordable price launched to the end user and the mobile application developers.  Inevitably, accessibility of such kind of device which has been used for years in the research and laboratory environments garnered public attention.  
	
	In this article, we introduce a signal pattern classification method in spectral domain and training procedures to classify the EMG signals obtained by 8-channel MYO armband real-time on a continuous manner for hand gesture and posture recognition.  The classification method is based on the Collaborative Representation based Classification (CRC) \cite{zhang2011sparse} which competes with the Sparse Representation based Classification (SRC) \cite{wright2009robust} with the same accuracy levels but much faster computation times. The drawbacks of the CRC compared to the SRC is that the CRC requires the observed and training signal patterns in the dictionary to be of equal length. This restricts the use of CRC methods for gesture and posture recognition study in which the duration of the gestures might vary for each repetition.  On the other hand, the SRC methods promotes the representation fidelity whereas the CRC data fidelity which might give rise to low classification accuracy, if the data do not exhibit fidelity \cite{zhang2012collaborative}. As the EMG signals are the products of complex stochastic processes, the CRC method yields poor recognition accuracy as we detail comparisons in the proceeding sections. 
	
	The Spectral Collaborative Representation based Classification (SCRC) proposed in this paper, overcomes these drawbacks and yield high recognition accuracy for a fairly rich hand gesture set.  As the observed and training signal patterns must have equal lengths, a special training scheme is adapted which allows building a training dictionary  with the representative columns  for all possible observed signal patterns. Therefore, the need to spotting signal patterns on a streaming signal is automatically eliminated, and the boundary of the representative columns are implicitly coded in the training matrix. 
	
	The contribution of this study to the literature are;
	
	\begin{itemize}
		\item The spectral content; complex conjugate eigenvalue pairs are obtained from 1D vectors by converting the observed signal pattern to a trajectory or a circulant matrix to capture the representative features of the EMG signals. 
		
		\item The gesture and posture recognition is performed in a continuous manner eliminating the requirements for spotting or picking the signal patterns on the streaming signals due to the proposed training scheme which implicitly embeds the signal boundaries on the representative columns.
		
		\item The training phase is easy to implement thus the end user can obtain a training dictionary on the spot. The flexibility in building a training dictionary paves the way for the use of the procedures and methods introduced here to implement different applications, such as control of a bionic prosthesis.  
		
		\item The number of hand gestures are the highest among the similar studies. Our worst recognition result is over 97\% with a substantial number of gesture labeling computation. 
		
		The rest of the paper is organized as follows. In Section 2, a brief review of the CRC method is given then the circulant matrix approach is detailed. In Section 3, after giving the technical details of the MYO armband, training phase for the gestures and SCRC is eloborated. The experiment and simulation results are discussed in Section 4. The paper ends with conclusion and future works in Section 5. 
		
	\end{itemize}
	
	\section{Spectral Collaborative Representation Based Classification}
	\subsection{Collaborative Representation based Classification}
	
	The spectral analysis which have been used to for decades in the studies where random signal or stochastic processes are involved. Although randomness in the systems might render prediction of some irregular features in the time domain, the spectral features such as eigenvalues and frequency might reveal valuable information of the underlying processes \cite{manolakis2005statistical}. The fourier analysis is the widely used spectral analysis method for this purpose. In this study, as the EMG signals show randomness, we exploit the spectral analysis by employing the circulant matrix structure for eigenvalue decomposition. 
	
	Both methods, the SRC and CRC address finding the linear representation coefficients vector $x$ from the linear systems of equation $y=Ax$ where  $A=[A_1,A_2,\ldots,A_n]  \in \mathbb{R}^{mxn}$  is the dictionary matrix in which the representatives from different classes are stacked as a column vector and $y$ is the observed signal. In the objective function of the solution (Eq. \ref{eqn:1}) different vector norms are utilized depending on the methods and requirements. 
	
	\begin{equation}
		\hat{x}=\argmin_x{\Arrowvert y-Ax\Arrowvert_{p}+\sigma\Arrowvert x \Arrowvert_q}
		\label{eqn:1}
	\end{equation} 
	
	In SRC solution $\ell_1$ regularization are used (p=2, q=1), where as the $p$ and $q$ becomes $1$ or $2$ in the CRC methods depending on the requirements such robustness of the classifier. When the objective function is solved by Regularized Least Square (RLS) in which $\ell_2$ norm used for both terms of the objective function the solution turns out the ridge regression. We used the least square version of the CRC method which is called as CRC\_RLS by the authors in \cite{zhang2012collaborative}. The ridge regression solution of Eq. \ref{eqn:1} thus is obtained as $\hat{x}=Py$ where $P   =({A^T}A+{\sigma}I)^{-1}{A^T}$ and $\sigma$ is the regularization parameter. Once the solution vector $\hat{x}$ is obtained the label of the observed signal is computed evaluating the minimum representation residuals $r_i$ given in Eq. \ref{eqn:2} where $\delta_{i}:\mathbb{R}^{n}\rightarrow\mathbb{R}^{n}$ is the selection operator that selects the coefficients of $i^{th}$ class while keeping other coefficients zero in the solution vector $\hat{x}$.    
	
	\begin{equation}
		\min_{i}\quad r_{i}(y)=\Arrowvert y-A\delta_{i}(\hat{x})\Arrowvert_{2}
		\label{eqn:2}
	\end{equation}
	
	The SRC methods yield good classification results as long as the dictionary matrix is overcomplete whereas the CRC do not require overcomplete dictionaries. In addition with the lower number of dictionary representatives, the regression operator $P$ is only computed and stored once and is coded over the anew observed patterns for the solution vector $x$ which makes the classification method perform faster then the SRC in real time applications. 
	
	Infact, even though new approaches have been introduced in the literature such as Block Sparse or Structured Sparse $\ell_1$ solvers\cite{zhang2012recovery, huang2011learning} and Block SRC methods perform much better then the plain SRC in terms of the computational time without compromising the accuracy levels \cite{boyali2014block}, the computation times are not satisfactory on the mobile devices such as phones and tablets for classification due to the fact that all the solvers are based on iterative algorithms \cite{boyali2015signal}. 
	
	\subsection{Circulant Matrices and Spectral Collaboration based Classification}
	
	Circulant matrices which appear naturally in many systems of equation such as that of dynamical systems, convolution and vibration theories \cite{manolakis2005statistical, lawrancebook, karner2003spectral}. A circulant matrix is constructed by its first row by shifting the elements of the row vector in clockwise or counterclockwise directions circularly. Depending of the direction of their diagonals (right or left circulants) they take the form of Toeplitz and Hankel matrices respectively which have wide range of application areas from compressing sensing, filter design, singular spectrum analysis to inverse eigenvalue problems \cite{bini1995toeplitz, henriques2015high, hassani2007singular, saad1992numerical}. 
	
	Let assume a vector $a=[v_0, v_1, \ldots, v_n]$, the right circulant $C =circ(a)\in\mathbb{R}^{nxn}$ becomes;
	
	\begin{equation}
		C_{a} =
		\begin{pmatrix}
			v_{0} & v_{1} & v_{2} & \cdots & v_{n-1} \\
			v_{n-1} & v_{0} & v_{1} &\cdots & v_{n-2} \\
			v_{n-2} & v_{n-1} &v_{0} &\cdots & v_{n-3}\\
			\vdots  & \vdots &\vdots  & \ddots & \vdots \\
			v_{1} & v_{2} &v_{3} &\cdots & v_{0}
		\end{pmatrix}
		\label{Eq:circm}
	\end{equation}

	Any circulant can be diagonalized by a unitary or Discrete Fourier Transformation (DFT) matrix due to the their cyclic structure. This factorization is also a similarity transformation. If any two matrices A and B are similar, one of the them can be expressed as  $B=T^{-1}AT$ where $T$ is an invertible matrix. If $T$ is a DFT matrix  $F_n$ (Eq. \ref{Eq:unitary} and Eq. \ref{Eq:facto}), the matrix $A$ becomes a diagonal matrix, the entries of which are the eigenvalues of the circulant transformed.

	\begin{equation}
		\Lambda=F_n^HCF_n
		\label{Eq:facto}
	\end{equation}
	
	In these equation  $diag(\Lambda)=\{\lambda_1, \lambda_2, \ldots, \lambda_n \}$ and the matrix operator $(o)^H$ is the Hermitian transpose.  
	
	\begin{equation}
		F_n =\frac{1}{\sqrt{n}}
		\begin{pmatrix}
			1 & 1 & 1 & \cdots & 1\\
			1 & W & W^2 & \cdots & W^{N-1} \\
			1 &  W^2 &  W^4& \cdots & W^2{N-2} \\
			\vdots  & \vdots  & \ddots & \vdots  \\
			1 & W^{N-1} & W^{2(N-2)} & \cdots &  W^{(N-1)(N-1)}
		\end{pmatrix}\
		\label{Eq:unitary}
	\end{equation}

	The columns of the DFT matrix $F_n$ with the entries $W=\exp{\frac{-2{\pi}i}{n}}$ thus becomes the eigenvectors of the circulant matrix. As all the eigenvectors of the decomposed circular matrices with the same size are all the same, the only discriminative subspace are only defined by the eigenvalues of the matrices.  These subspaces which are shift invariant can be used as a dictionary matrix in the representation based classifications.  The time complexity of the eigenvalue decomposition with a number of operation $\mathcal{O}(n\log{}n)$ is lower then the conventional decomposition methods, therefore, eigenvalue decomposition of the circulants is faster then the conventional eigenvalue decomposition algorithms \cite{henriques2015high}.  
	
	In the SCRC, we build a training dictionary using the eigenvalues of the observed signal as the features. Since, complex numbers are involved, the Hermitian transpose of the vectors is used in the inner product space. In the complex domain, the solution vector $\hat{x}$ is obtained by the equation $\hat{x}=P_cx$ where;
	
	\begin{equation} 
		P_c=({A^H}A+{\sigma}I)^{-1}{A^H}  
	\end{equation}
	
	The training matrix is normalized and centered in the CRC methods while the observed signal $y$ is only centered. It is important to note that since the spectral features are used in the Fourier domain, Hermitian conjugate transpose is used for all inner product operations.  
	
	\section{SCRC Application in EMG signal Classification}
	\subsection{MYO Armband and Gestures}
	
\begin{figure}[h]
		\centering
		\includegraphics[scale=0.3, keepaspectratio=true]{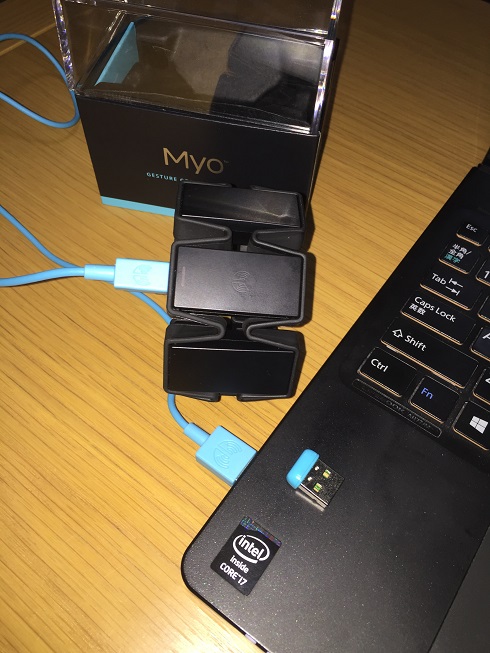} 
		\caption{MYO Armband Kit with Bluetooth Dongle}
		\label{fig:f1}
	\end{figure}

	The MYO armband which has been recently released to the market by Thalmic Labs revolutionized the EMG based studies in which previously the EMG signals were only acquired in the laboratory or research environment with the expensive hardware.  The sensor armband has 8-channel EMG sensor group along supported with an Inertial Measurement Unit (IMU) which reports linear and rotational acceleration as well as the rotation angles around 3-axes. The sensor readings are transfered over a low power bluetooth adapter to the computer (Fig. \ref*{fig:f1}).

	The Software Development Kit (SDK) allow the developers to access the EMG signals and motion parameters on the worn arm. As the measured EMG signals depend on the sensor location on the muscles, the MYO armband applications require a special calibration gesture must be performed every time when the armband is put on the arm or taken off.  In this way, the application calibrate the sensor locations depending on the sensed sensor direction and rotation of the armband. In our experiments we collected the experimental data without giving a break and taking off without needing re-calibrating the device. 
	
\begin{figure}[h]
		\centering
		\includegraphics[width=\columnwidth, keepaspectratio=true]{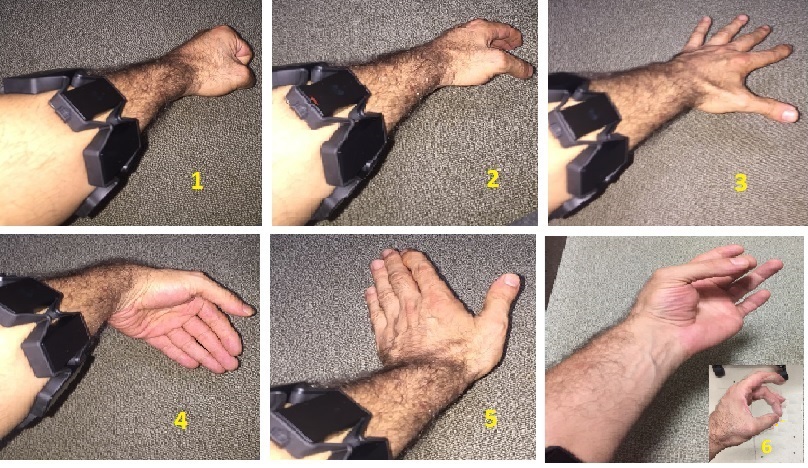} 
		\caption{MYO Hand Gestures, 1- Fist, 2- Hand Relax, 3- Finger Spread, 4- Wave In, 5- Wave Out, 6- Double Tap}
		\label{fig:f2}
	\end{figure}

	The IMU unit reports the motion related parameters at a frequency of 50 Hz, whereas the frequency for the EMG signal is 200 Hz as reported by the company website \cite{myoweb}. The SDK provides a gesture object by which five different hand gestures are reported. The same hand gesture set which is composed of Hand Fist, Wave In, Wave Out, Hand Spread and Double Tap is used to test the SCRC performance (Fig. \ref{fig:f2}). If none of them is performed, the hand is assigned to the hand relax gesture.  
	
	The location of the EMG sensors are important for certain types of gestures as different hand and finger movements are controlled by the different muscle groups on the forearm. The armband configuration are put on close to the elbow where the discriminant EMG signals can be captured for hand movements. In order to classify individual finger gestures, the EMG sensors must be placed on the related muscle groups for finger resolution.

	\subsection{Training}
	
	In gesture recognition studies one of the open problems is spotting the gestures on a streaming signals. A gesture has a start and end points which can be spotted on a continuously monitored signal, however gestures might differ among the people even for the same performer. Along with the intra and inter personal variations, the time length of the different gestures in a set might also show variations. The spotting problem manifests itself in building a training dictionary for the representation based classification methods. This problem can be shortcut using a switch which is used to mark the beginning and end points of the gestures in real world and a trigger in the software applications. In the study, the authors collect gestures by drawing in the air using an IR pen which is tracked by the IR cameras of a stereo vision Wiimote experimental setup. The performer press and release the switch of the IR pen at the beginning and the end of the gestures respectively. Spotting can also be done visually by analyzing the trajectory or the visually discriminating features which requires tedious effort for fairly rich gesture sets.
	
	In this study, as the EMG signals are highly jittery and the gesture patterns cannot be spotted visually on the figures of the signal we employ subspace clustering methods for the repeated couple gestures. We used this approach in our previous studies. The reason is twofold to adapt this approach. The first is the drawback of the constant sliding window which is also important for the CRC method. Both in the real-time applications and the training phase for building a representative dictionary, the signals are captured by a sliding window. The second is finding representatives for every gesture state in any time window which affect the recognition accuracy. As the tracked body part, in our case the hand might switch between gestures while being tracked, in the switching regions there occur overlapping gestures in the monitoring time window which degenerates the robustness and accuracy of the recognition. 
	
	We overcome these two drawbacks employ a subspace method for a repeated gesture couple. At least two gestures are necessary to cluster them into respective classes for the subspace clustering algorithms, the state of art exploits the self similarity property of the signals to be clustered. The most eminent of these algorithms are the Sparse and Low Rank Subspace Clustering (SSC, LRR) methods which make use of $\ell_1$ and nuclear norm in their objective functions \cite{elhamifar2013sparse, liu2013robust}. We use a variant of SSC, the Ordered Subspace Clustering (OSC) which puts an additional penalty in the self similarity objective function aiming the sequential data \cite{tierney2014subspace}. The additional term in the objective function enforces the neighboring representative samples to be same by penalizing the differences of two neighbors.   
	
\begin{figure}[H]
		\centering
		\includegraphics[width=\columnwidth, keepaspectratio=true]{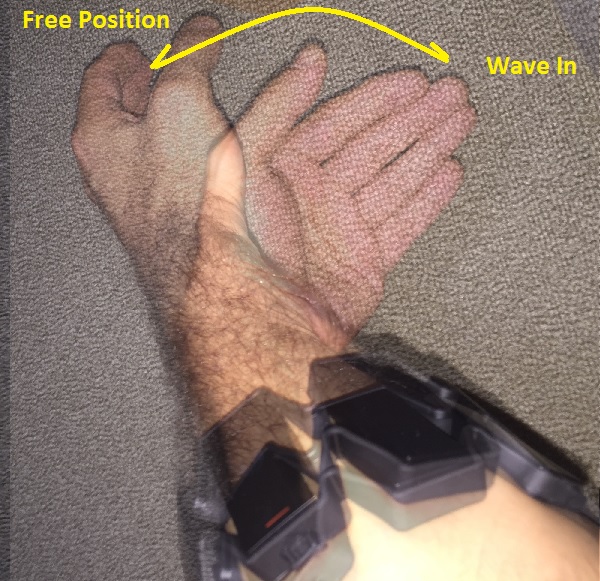} 
		\caption{Wave In - Hand Relax - Wave In Repetition}
		\label{fig:f3}
	\end{figure}

	We collected gesture couples for clustering the obtained signal into two different classes using the OSC. The hand performs two gestures repeatedly. As the hand the hand relax state (which is actually not a gesture) every gestures ends with hand relax state. This state is assumed to be gesture mate of the each gesture. As a concrete explanation, let's assume that a hand performs the hand wave in gesture and returns to the hand relax position repeatedly as in demonstrated in Fig. \ref{fig:f3}.
	
	On the recorded EMG signal, a constant length sliding window with an increment of sampling interval capture the content which is then put into the combined dictionary matrix. The subspace clustering method is used in this step to distill the pure representatives into their respective classes.  A sample clustering result is given in Fig. \ref{fig:f3} for the double tap gesture on a single channel EMG signals. As seen, the period of the one gesture cycle is approximately 0.5 seconds corresponding to 100 time samples.
	
	\begin{figure}[H]
		\centering
		\includegraphics[width=\columnwidth, keepaspectratio=true]{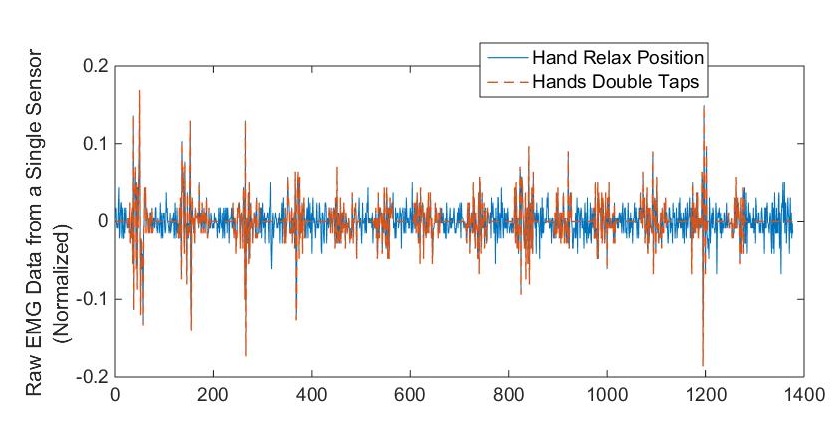} 
		\caption{Clustered Double Tap - Hand Relax Gestures on EMG data}
		\label{fig:f4}
	\end{figure}
	
	In fact, although it is not seen on the figure,  there are four hand states in repetitions of the gesture couples; two static hand postures where the hand waits at the hand relax and terminal waved in positions, and the dynamics hand gestures through these postures. However, as the number of subspaces are chosen as two for the gesture couples, the posture states are embedded into the clusters automatically and every observable hand states are represented in these clusters. The training dictionary contains 10 gesture classes in this case and we map every return states to the hand relax position from any hand gesture are mapped to the five hand gestures. 
	
	\section{Simulation Results} 
	
	We collected five gesture couple experiments, one for training and the rest for testing. In addition, we perform an arbitrary hand gesture sequence in which the hand performs arbitrarily performs gestures arbitrarily in the sequence. We also repeated the recognition simulations using the conventional CRC\_RLS and compared the results.
	
\begin{figure}[H]
		\centering
		\includegraphics[width=\columnwidth, keepaspectratio=true]{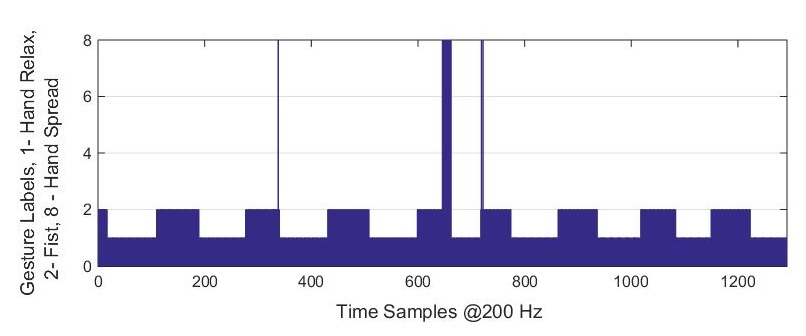} 
		\caption{Wave In - Hand Relax Gestures on EMG data with CRC}
		\label{fig:f5}
	\end{figure}

	Although, the conventional CRC\_RLS yields the same recognition accuracy for some of the experimental data, it can not discriminate the hand wave in and fist gestures labeling the wave in as fist in all the experiments (Fig. \ref{fig:f5}) with some number of misclassification with the hand spread gesture.  However, the worst recognition result for the hand wave in gestures among the four test sets is 99.36\% with the SCRC method. 
	
\begin{figure}[H]
		\centering
		\includegraphics[width=\columnwidth, keepaspectratio=true]{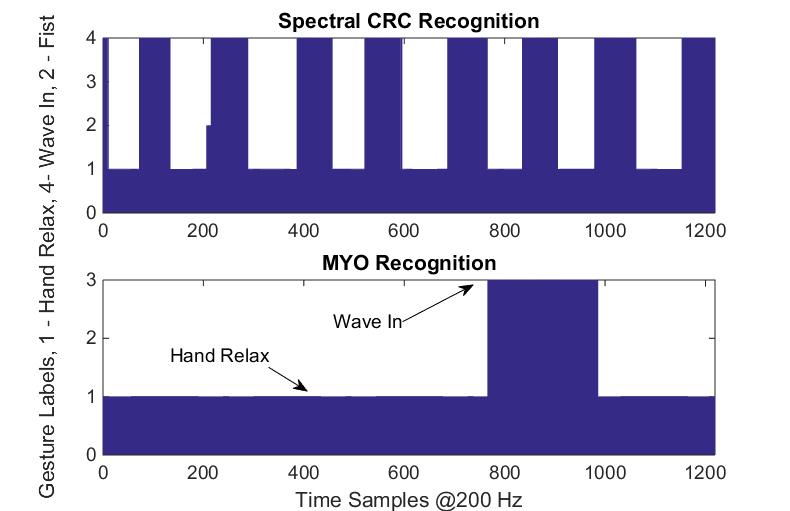} 
		\caption{Wave In - Hand Relax Gestures on EMG data with SCRC}
		\label{fig:f6}
	\end{figure}

	Fig. \ref{fig:f6} shows the worst result of the wave in - hand relax test experiment. The bottom figure is the gesture label of the MYO gesture recognition module. Since the SRC recognizes 10 gestures the number of labels are different then the MYO gesture module which labels the gestures as 1- Hand Relax, 2- Fist, 3- Wave In, 4- Wave Out, 5- Hand Spread and 6- Double Tap. The number of labeling computation in the figure is 1219 and the SCRC mis-classifies the signal patterns as the fist gesture eight times.  
	
\begin{figure}[H]
		\centering
		\includegraphics[width=\columnwidth, keepaspectratio=true]{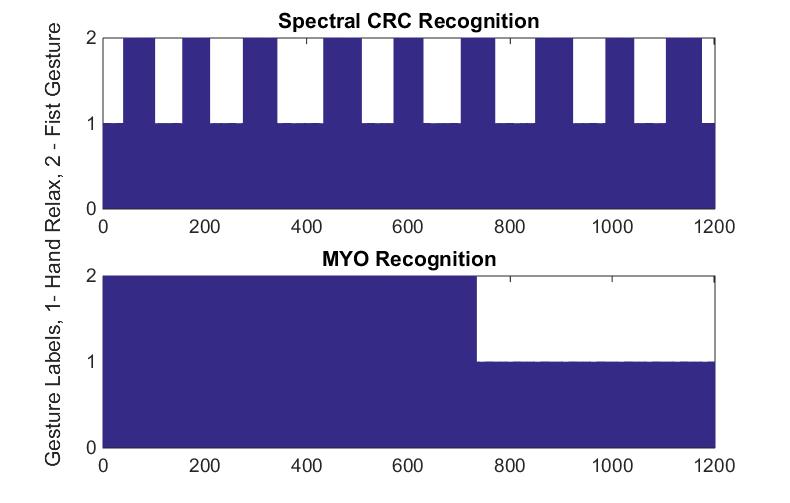} 
		\caption{Recognition Results for Fist and Hand Relax Experiment, Spectral CRC and MYO}
		\label{fig:f7}
	\end{figure}

\begin{figure}[H]
		\centering
		\includegraphics[width=\columnwidth, keepaspectratio=true]{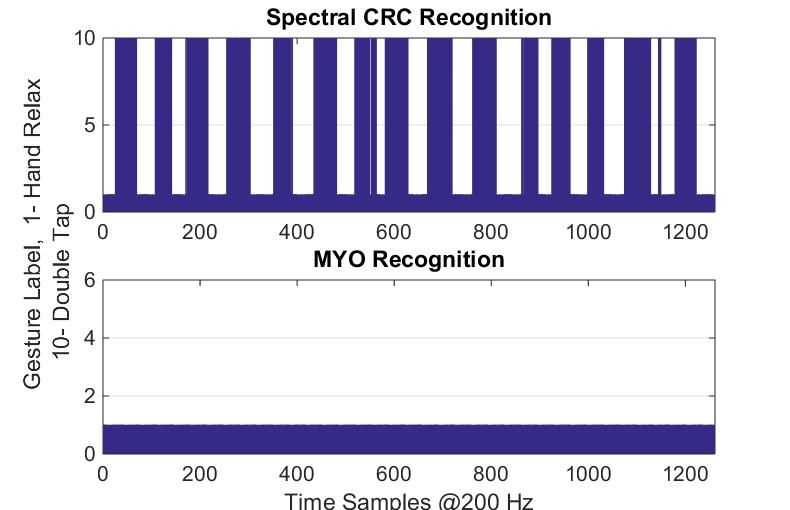} 
		\caption{Recognition Results for Double Tap and Hand Relax, Spectral CRC and MYO}
		\label{fig:f8}
	\end{figure}

	The recognition accuracy the fist and the double-tap gestures is 100\% for all the test experiments (Figs. \ref{fig:f7}-\ref{fig:f8}).
	
\begin{figure}[H]
		\centering
		\includegraphics[width=\columnwidth, keepaspectratio=true]{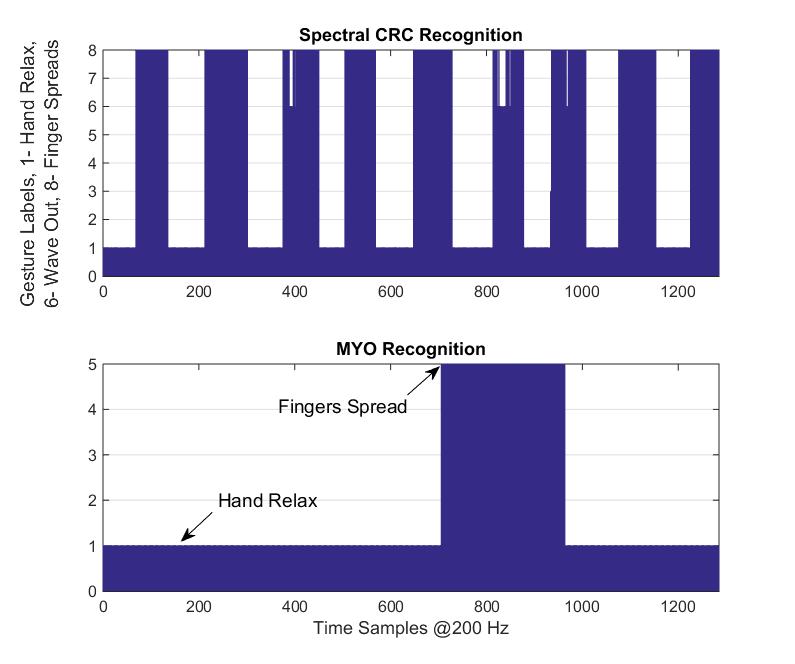} 
		\caption{Recognition Results for Hand Spread and Hand Relax, Spectral CRC and MYO}
		\label{fig:f9}
	\end{figure}

	The results of the fingers spread recognition is given in Fig. \ref{fig:f9}. The worst recognition accuracy in the study among all the experiment is obtained for this hand gesture as 97.3\%. The algorithm gives some mis-classifications for   finger spreads and the wave out couple as there are overlapping wrist contraction and flexion states for the two hand movements that the hand might visit involuntarily.  
	
\begin{figure}[H]
		\centering
		\includegraphics[width=\columnwidth, keepaspectratio=true]{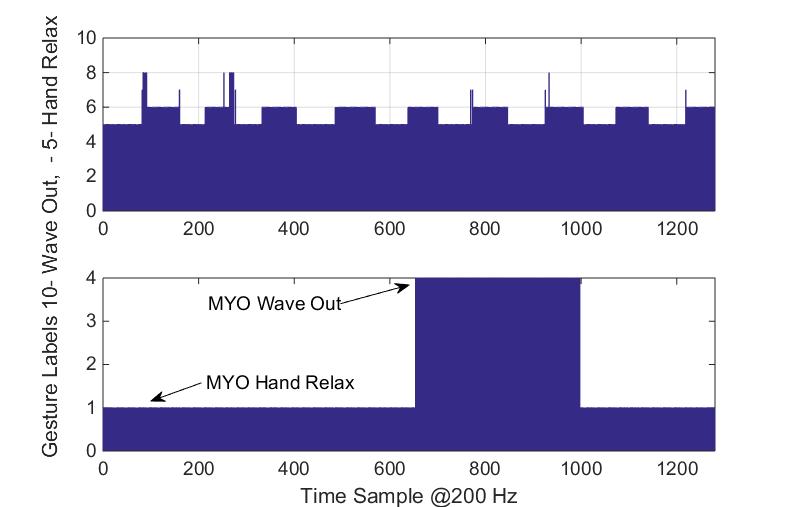} 
		\caption{Recognition Results for Wave Out and Hand Relax, Spectral CRC and MYO}
		\label{fig:f10}
	\end{figure}

\begin{figure}[H]
		\centering
		\includegraphics[width=\columnwidth, keepaspectratio=true]{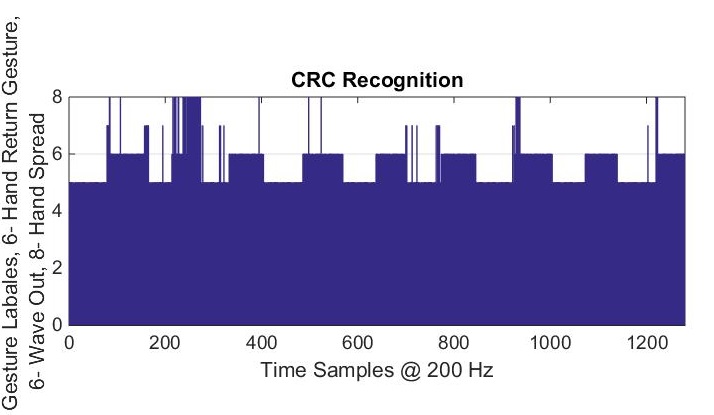} 
		\caption{Recognition Results for Wave Out with CRC\_RLS }
		\label{fig:f11}
	\end{figure}

	The hand wave out gesture recognition results are given in Fig. \ref{fig:f10} without mapping the hand relax return gesture to the hand relax state. In all the figures the mapped hand relax state is labeled as one. In the following figure the label six show that the hand is moving toward to the hand relax position. The worst recognition accuracy out of four test experiments is 98.34\% for 1274 labeling computation in the wave out experiment. The results when the CRC\_RLS is given in Fig. \ref{fig:f11}, in which the accuracy is less then 85\%. 
	
\begin{figure}[h]
		\centering
		\includegraphics[width=\columnwidth, keepaspectratio=true]{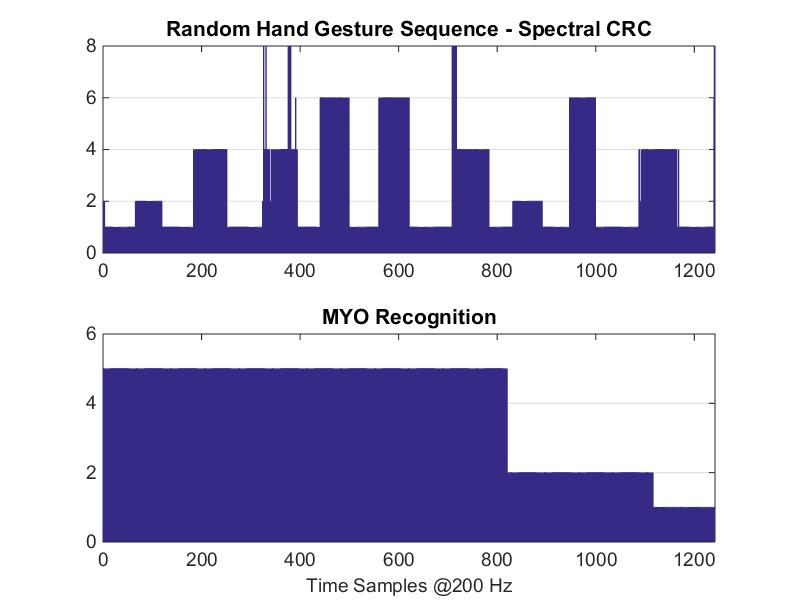} 
		\caption{Recognition Results for Random Hand Gesture Sequence, Spectral CRC and MYO, 1- Hand Relax, 2- Fist, 4- Wave In, 6- Wave Out, 8- Hand Spread}
		\label{fig:f12}
	\end{figure}

	As the final experiment we give the results of an arbitrary gesture sequence. In this experiment, unlike the other experiments where the hand only performs two hand gestures, different gestures are performed arbitrarily (Fig. \ref{fig:f12}). The recognition accuracy is 98.47\% for this experiment. 
	
	\section{Conclusion and Future Works}
	
	In this study we introduced a novel application of the CRC methods in Fourier domain and elaborated the methodologies and experimental results for the hand gesture recognition using the EMG measurements on a forearm. The accuracy levels for the fairly rich gesture set is promising for the raw EMG signals, therefore, they can be further improved by analyzing the signal noise and introducing the appropriate filtering methods. The use of Fourier features is explained intuitively using the circulant approach which reveals shift invariant features in an elegant way. The approach detailed in the training phase which we used for detecting braking maneuvers of a mobility robot first time \cite{boyali2015paradigm} leads to a dictionary with a high power of discrimination. The embedding of the gesture boundaries into the representative states are implicitly realized by the subspace clustering methods.   
	
	The methods and procedures we detailed in this paper is a small portion of a project by which a multi-modal intuitive HCI have been developed for the elderly or severely handicapped people.  The motivation of the project to enable these people to steer a robotic wheelchair without excessive cognitive or physical efforts. On the other hand the developed systems will be employed for augmented and virtual reality environments to establish training theaters for people who are prescribed power wheelchairs first time. 
	
	\section*{Acknowledgments}
	
	The study is supported by the Japan Society for the Promotion of Science (JSPS) fellowship program and the KAKENHI Grant (Grant Number 15F13739).  
	
	\section*{References}
	
	\bibliography{mybibfile}
	
\end{document}